\documentclass[lettersize,journal]{IEEEtran}
\usepackage{amsmath,amsfonts}
\usepackage{algorithm}
\usepackage{array}
\usepackage[caption=false,font=normalsize,labelfont=sf,textfont=sf]{subfig}
\usepackage{textcomp}
\usepackage{stfloats}
\usepackage{url}
\usepackage{verbatim}
\usepackage{graphicx}
\usepackage{cite}
\usepackage{epsfig}
\usepackage{graphicx}
\usepackage{graphics}
\usepackage{amssymb}
\usepackage{bbding}
\usepackage{algpseudocode}
\usepackage{comment}
\usepackage{url} 
\usepackage{amsmath,amssymb} 
\usepackage{color}
\usepackage{algorithm}
\usepackage{algpseudocode}
\usepackage{threeparttable}
\usepackage{multirow}
\usepackage{makecell}
\usepackage{booktabs}
\usepackage{bm}
\usepackage{array}
\usepackage{algorithm}
\usepackage{algpseudocode}
\usepackage[pagebackref=true,breaklinks=true,colorlinks,bookmarks=false]{hyperref}
\usepackage{epsfig}
\usepackage{bbding}
\usepackage{cite}
\usepackage[T1]{fontenc}
\usepackage[shortcuts]{extdash}

\usepackage{adjustbox} 
\usepackage{pifont}
\newcommand{\cmark}{\ding{51}}
\newcommand{\xmark}{\ding{55}}
\usepackage{tabularx}
\usepackage[normalem]{ulem}
\usepackage{algorithm}
\usepackage{algpseudocode}
\usepackage{hyperref}
\usepackage{verbatim}
\usepackage{setspace}
\usepackage[pagebackref=true,breaklinks=true,colorlinks,bookmarks=false]{hyperref}
\usepackage{amsthm,amsmath,amssymb}\usepackage{mathrsfs}
\usepackage{svg}
\usepackage{array}
\usepackage{booktabs} 

\newcolumntype{L}[1]{>{\raggedright\arraybackslash}p{#1}} 
\newcolumntype{C}[1]{>{\centering\arraybackslash}p{#1}}   
\newcolumntype{R}[1]{>{\raggedleft\arraybackslash}p{#1}}  

\begin{document}

\title{PixelSR: Efficient Screen Content Super-Resolution \\ via Pixel Classification}

\author{Zhiheng Li, Lei Chen,~\IEEEmembership{Member,~IEEE},
Jie~Zhou,~\IEEEmembership{Fellow,~IEEE}, and Jiwen~Lu,~\IEEEmembership{Fellow,~IEEE}
\thanks{
This work was supported in part by the National Key Research and Development Program of China under Grant 2023YFB280690, and in part by the National Natural Science Foundation of China under Grant 62321005, Grant 62336004, and Grant 62125603.
\emph{(Corresponding author: Lei Chen)}
The authors are with the Department of Automation, Tsinghua University, Beijing, 100084, China.
\emph{(E-mail: 
lizhihan21@mails.tsinghua.edu.cn;
leichenthu@tsinghua.edu.cn; 
jzhou @tsinghua.edu.cn;
lujiwen@tsinghua.edu.cn.)}
}
}

\markboth{Journal of \LaTeX\ Class Files,~Vol.~14, No.~8, August~2021}%
{Shell \MakeLowercase{\textit{et al.}}: A Sample Article Using IEEEtran.cls for IEEE Journals}


\maketitle

\begin{abstract}
Screen content images are generally composed of texts and graphics. Compared to natural images, these man-made images contain a large quantity of sharp but repetitive structures. However, existing works in screen content super-resolution underutilize the special characteristics of screen content, leaving a large room to improve model performance and speed up. In this paper, we propose PixelSR, a simple yet effective method to improve super-resolution performance but with faster inference speed. To improve model performance, we classify pixels via pixel binning to compute content attention in the training phase. Specifically, after binning pixels into content-dependent groups, content attention is aggregated from pixel features within each group to introduce a content-dependent and non-local receptive field for every pixel. In the testing phase, we utilize the properties of self-repetitiveness and redundancy in screen content to speed up inference without the loss of model performance. We divide targeted high-resolution pixels into three types, which are unique pixels, repeated pixels, and background pixels for each test image. We conduct conventional network processing on unique pixels and cache their predictions in the on-the-fly lookup table. For repeated pixels which have appeared in unique pixels, we directly retrieve prediction results from the lookup table without network processing. For background pixels, we use the nearest neighbor algorithm to generate high-resolution pixels. The on-the-fly lookup table is cleaned and repeats the procedure above for the next test image. Experiments show our PixelSR achieves state-of-the-art performance with shorter inference time in screen content super-resolution.
\end{abstract}


\section{Introduction}
\label{sec:intro}
 Screen content images have become widespread with the development of multimedia applications. However, due to the limited bandwidth in wireless transmission, screen content images are likely to be transmitted with low resolutions (LR). Thus, efficient screen content SR is important for real-time high-resolution (HR) display. Since the properties of screen content images are significantly different from natural images \cite{sr1,sr2, sr3, sr4, sr5,sr6,sr7,sr8,sr9,sr10,sr11,sr12,sr13,sr14,sr15,anycost,sr16,sr17,sr18,sr19,sr20,sr21,sr22,ttsr,masa}, conventional SR of natural images do not generalize to screen content images \cite{btc,yang2015perceptual}. 

Efficient SR has been widely investigated in natural images. As shown in Fig. \ref{fig1}(a), previous efficient SR works \cite{kong2021classsr, wang2024camixersr} generally conduct sub-image classification and process easy and hard windows by separate networks. Similarly, PCSR\cite{PCSR} in scale arbitrary SR also classifies pixels into easy and hard, and processes them by a simple and complex network, respectively. These works achieve faster inference speed but with a performance drop. As for screen content SR, existing papers \cite{yang2015perceptual, btc} demonstrate the distribution discrepancy between natural and screen content images, showing conventional SR methods for natural images are not sufficient for the screen content case. To achieve efficient screen content SR, BTC \cite{btc} adopts B-splines basis functions to learn sharp fluctuations. However, the receptive field of BTC is limited in a local and small region. To expand the receptive field in SR, previous SR models \cite{ciaosr,crossscalenonlocal,gnn} conduct cosine similarity between pixel features to inject information from non-local features. However, the memory cost of this operation is ultrahigh, requiring window partitioning which slows down the SR process a lot. In addition, it did not fully utilize the properties of screen content images for efficient SR. Screen content are man-made images, composed of sharp and repetitive structures for vector graphics. For example, although one article is long, it is generally made of 26 letters and several symbols with the same font style and font size on the screen. Thus, the contents of each screen content image frequently repeat themselves at the pixel level, and the edges of these texts are unnaturally sharp. Also, the background of screen content images consists of only one RGB value without any fluctuations. These properties rarely happen in natural images, and should not be missed for efficient screen content SR. 

\begin{figure}[t]
  \centering
  \label{fig1}
  \includegraphics[width=0.47\textwidth]{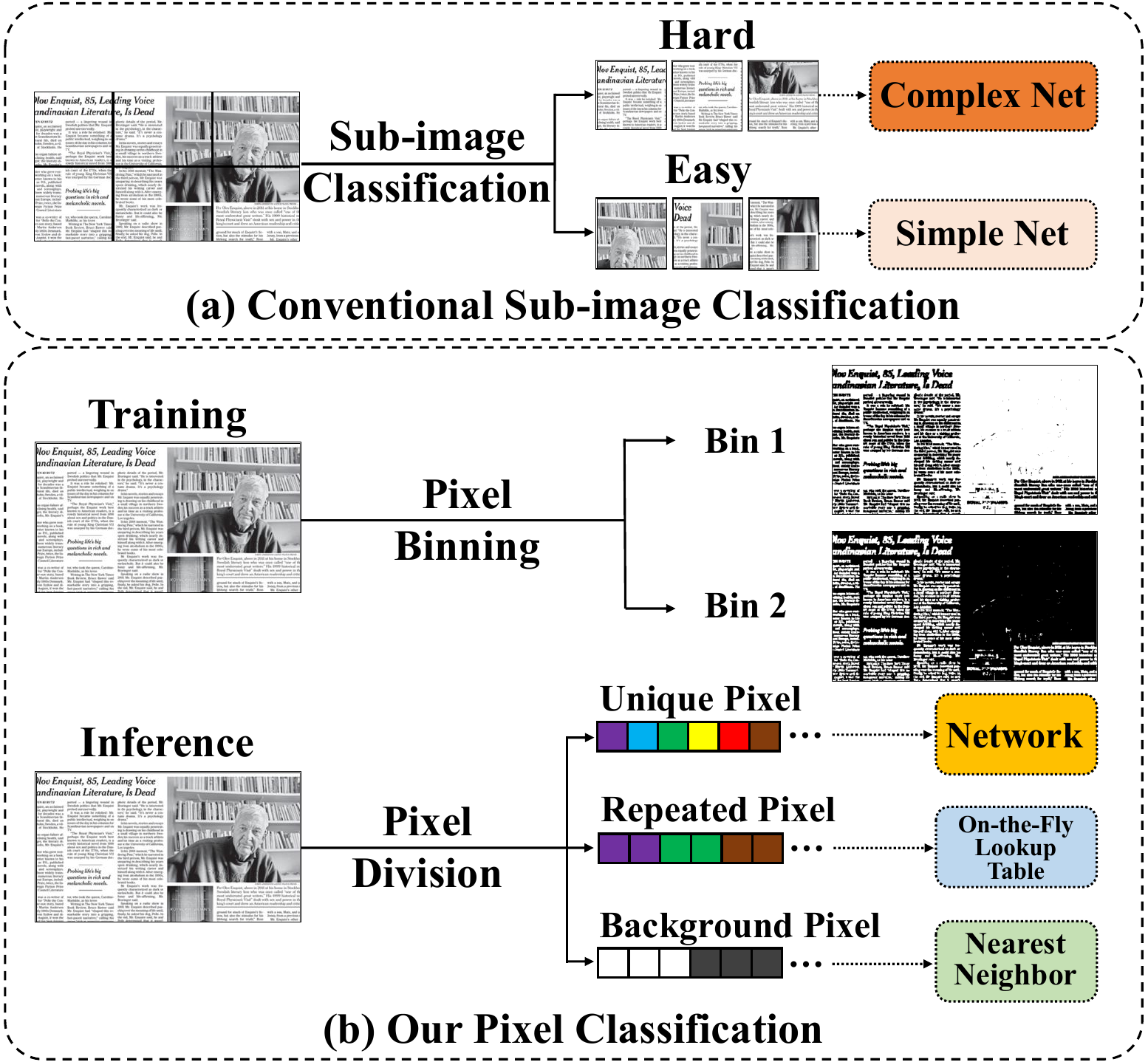} 
  \caption{\label{fig1} Comparison between PixelSR and conventional efficient SR paradigm: (a) Conventional efficient SR classifies sub-images into easy and hard, and processes them by separate networks. (b) PixelSR classifies image content at the pixel level. In training, PixelSR classifies pixels into content-dependent groups via pixel binning. By computing content attention from pixel groups, PixelSR assigns a content-dependent and non-local region for PSNR improvement. In inference, PixelSR divides targeted HR pixels into unique pixels, repeated pixels, and background pixels. After conventional network processing on unique pixels, PixelSR predicts repeated pixels and background pixels by the on-the-fly lookup table and the nearest neighbor algorithm respectively, with faster inference speed but no PSNR drop. }
\end{figure}

In this paper, we propose pixel classification to improve the training and inference for efficient screen content SR. In the training phase, we particularly utilize the property of unnatural sharp edges of screen content to improve model performance. Specifically, we introduce the content extractor with binning to split pixels into two content-dependent groups, where one group is composed of sharp edges of screen content and another group consists of the rest. However, hard binning breaks backpropagation and stops the gradient flow. Therefore, we design the soft binning technique to ensure the training of the content extractor. Unlike the correlation operation between dense pixel features which is used in previous works \cite{ciaosr,crossscalenonlocal,gnn} to group similar pixels but is computationally expensive and inference slow, the content extractor with soft binning discriminates the edges of screen content from other regions over the whole image with little time and memory addition. In addition, the number and locations of pixels in each group are various. To learn over scattered pixels, we merge them by using pooling operations to construct content attention and concatenate them to every pixel before feeding into the implicit neural representation. In this way, we assign a content-dependent and unlimited receptive field to each pixel, which efficiently improves PSNR. As shown in Fig. \ref{fig1}(b), the proposed pixel binning splits pixels into two groups, and roughly separates them based on edges and plain pixels. 

In addition to training, we design an on-the-fly lookup table to speed up inference without PSNR loss. We divide HR targeted pixels into three types, which are unique pixels, repeated pixels, and background pixels, as shown in Fig. \ref{fig1}(b). For the background pixels, we replace the complex network processing with the simplest nearest neighbour algorithm to generate HR pixels. Excluding background pixels, the rest pixels are unique pixels and repeated pixels. Repeated pixels are the pixels which have occurred in unique pixels. We use conventional network processing to predict RGB values of unique pixels, and cache these RGB prediction results in the on-the-fly lookup table. For repeated pixels, we directly retrieve the RGB outputs from the lookup table without the network processing. Since the on-the-fly lookup table utilizes the property of self-repetitiveness of screen content images, we clean the cached lookup table after finishing the inference of one test image and then repeat the procedure above. Therefore, we generate a unique lookup table for every test image. We will show the proposed on-the-fly lookup table can be adopted in previous methods such as LIIF\cite{liif}, ITSRN\cite{itsrn}, LTE\cite{lte}, and BTC\cite{btc} to speed up inference without performance drop. 


In summary, the contributions of PixelSR are threefold. First, in training, we classify pixels into content-dependent groups via the content extractor with pixel binning. By aggregating pixel features of each group, we introduce content attention and assign content-dependent and non-local information to each pixel. Second, we divide HR targeted pixels into unique pixels, repeated pixels, and background pixels in inference. We design an on-the-fly lookup table and use the nearest neighbor algorithm to speed up the inference of repeated pixels and background pixels without PSNR loss. Third, PixelSR achieves state-of-the-art performance with shorter inference time compared to the baselines. 

\section{Related Work}
In this section, we review the related topics. First, we review the scale arbitrary SR, which upscales natural images by a random scale factor. Then, we discuss the screen content SR, which focuses on screen content images. Next, we review efficient fixed-scale SR of natural images. Finally, the research field of lookup table in SR is presented. 

\subsection{Scale Arbitrary SR}
The task of scale arbitrary SR is to upscale one LR image at any scale factor by using only one model. MetaSR \cite{metasr} first proposes this task, and introduces a meta-upscale module to predict the weights of the upscale filters dynamically. ArbSR \cite{arbsr} proposes the scale-aware adaptation and upsampling components to improve the model performance. Different from previous models which are based on CNN, LIIF \cite{liif} first uses the implicit neural representation to achieve the goal of scale arbitrary SR. It introduces a local implicit image function to reconstruct HR pixels at arbitrary coordinates and arbitrary scale. LTE \cite{lte} introduces a local texture estimator to reconstruct HR signals in the frequency domain. EQSR \cite{eqsr} designs an adaptive feature extractor to achieve scale-aware representation learning. For upsampling, a learnable neural kriging upsampling operator is used for scale-aware spatial feature fusion. Both techniques are used to boost model performance in scale arbitrary SR of natural images. UltraSR \cite{ultrasr} focuses on spatial encoding and thus designs positional encoding to enhance model performance. LINF \cite{linf} proposes a local implicit normalizing flow to model texture distribution via normalizing flow. CLIT \cite{clit} extracts multi-scale features with a cumulative strategy to gradually increase the upscale factor during training. \cite{actmore} introduces a scale-aware local feature adaptation module to adjust the dynamic filters based on the feature and scale. A local feature adaptation upsampling module is followed to reconstruct HR pixels at arbitrary scale. CiaoSR \cite{ciaosr} proposes a continuous implicit attention-in-attention network. Specifically, it uses an implicit attention network to learn the ensemble weights. Also, a scale-aware attention is designed to exploit non-local features. IPF \cite{ipf} introduces an implicit pixel flow to learn the offsets of the coordinates for photorealistic scale arbitrary SR. HIIF \cite{hiif} proposes a hierarchical implicit image function to capture features at multiple scales to facilitate reconstruction of HR pixels. Although the methods above achieve better performance in scale arbitrary SR, they do not aim to realize efficient scale arbitrary and only work for natural images. 

\subsection{Screen Content SR} 
SISR has been investigated in the sub-field of screen content images~\cite{hu2019meta,liif,arbsr}. Since screen contents are man-made images, their pixel distribution is distinct from that of natural images. \cite{yang2015perceptual} presents an analysis of screen content images. It presents the discrepancy between the natural and screen content images, validating SR of natural images does not generalize to screen content. \cite{yang2012learning} proposes a learning approach to construct a dictionary for screen image compression. \cite{wang2021super} designs an efficiency SR structure and a loss function to super-resolve compressed screen content videos effectively. 
ITSRN~\cite{itsrn} enhanced the quality of screen content images by leveraging implicit transformer-based relationships between LR and HR coordinates. 
Building on this, ITSRN++~\cite{shen2022itsrn++} introduced a cyclic modulated implicit transformer-based upsampler and enhanced transformer-based feature extraction to further improve SR performance for screen content images.
BTC method~\cite{btc} utilized non-uniform B-spline basis functions to represent image features and implicit neural representation to achieve superior performance.
Although it is effective in reconstructing sharp edges, we suggest previous methods did not fully explore the properties of screen content images for efficient SR. 

\subsection{Efficient SR of Natural Images.}
Recently, efficient SR has been developed to reduce computational requirements and redundant parameters~\cite{shi2016real,CARN, IMDN, EDSR-baseline, luo2022lattice, kong2021classsr,DLGSANet-tiny}.
Recently, there has been significant interest in developing efficient SR methods to alleviate computational demands and reduce redundant parameters. Various strategies have been proposed to enhance the efficiency of SR algorithms~\cite{dong2016accelerating, shi2016real, li2023ntire, luo2022lattice, wang2023global}.
For designing more efficient frameworks, EDSR-baseline~\cite{EDSR-baseline} optimizes the architecture by removing the batch normalization layer, resulting in a deep and wide network with residual blocks. ClassSR~\cite{kong2021classsr} classifies whether sub-images are processed by complex, medium, or simple networks, reducing 50\% computational costs on 8K datasets. CAMixerSR~\cite{wang2024camixersr} distinguish informative regions and plain areas by routing content-aware token mixer design. LMLT \cite{lmlt} introduces the low-to-high multi-level transformer. By utilizing attention with varying sizes for each head, LMLT achieves competitive results with lower inference time and GPU memory usage. PCSR \cite{PCSR} classifies pixels into simple and hard types, and uses simple and complex nets to reconstruct HR RGB values. It fastens inference speed but a performance drop is unavoidable. In contrast, our PixelSR achieves shorter inference time with better PSNR performance via pixel classification. 

\subsection{Lookup Table SR.}
Lookup table SR has emerged as a promising approach to balancing computational efficiency and performance. 
As a network construct, SR-LUT~\cite{jo2021practical} leverages lookup tables by training a simple deep SR network to convert inputs and outputs into a lookup table. To boost SR performance, Li et al.~\cite{li2022MuLUT} introduced MuLUT, which increases the receptive field by using hand-crafted indexing patterns and cascading lookup tables.
Ma et al.~\cite{ma2022SPLUT} proposed a series-parallel lookup table framework, introducing channel-level lookup tables and processing two components separated from the original 8-bit input in parallel. 
Both MuLUT and SPLUT divide the single lookup table of SR-LUT into multiple lookup tables, effectively increasing the receptive field of SR models. However, their enhanced performance comes at the cost of increased lookup table size, demanding substantial storage resources. The trade-off between receptive field size and lookup table size remains a significant challenge.
RCLUT~\cite{liu2023reconstructed} addresses the burden of large lookup table sizes when increasing the receptive field by employing a novel reconstructed convolution technique. This method approximates the performance of vanilla convolution with a drastically smaller lookup table size. 
By integrating reconstructed convolution, RCLUT achieves efficient SR with an expanded receptive field but without the exponential increase in lookup table size, offering a practical solution to the trade-off issue and paving the way for more efficient and scalable SR models.
As an accelerator plugin, ARM~\cite{chen2022arm} and CABM~\cite{tian2023cabm} design a strategy to build an Edge-to-Bit lookup table that maps the edge score of a patch to the bit of each layer during inference, when observe that the edge information can be an effective metric for the selected bit. To expand the receptive field and enhance information flow, AutoLUT \cite{autolut} introduces automatic sampling with pixel abstractions and the AdaRL layer to improve model performance. 


\begin{figure*}[t!]
  \centering
  \label{fig3}
  \includegraphics[width=\textwidth]{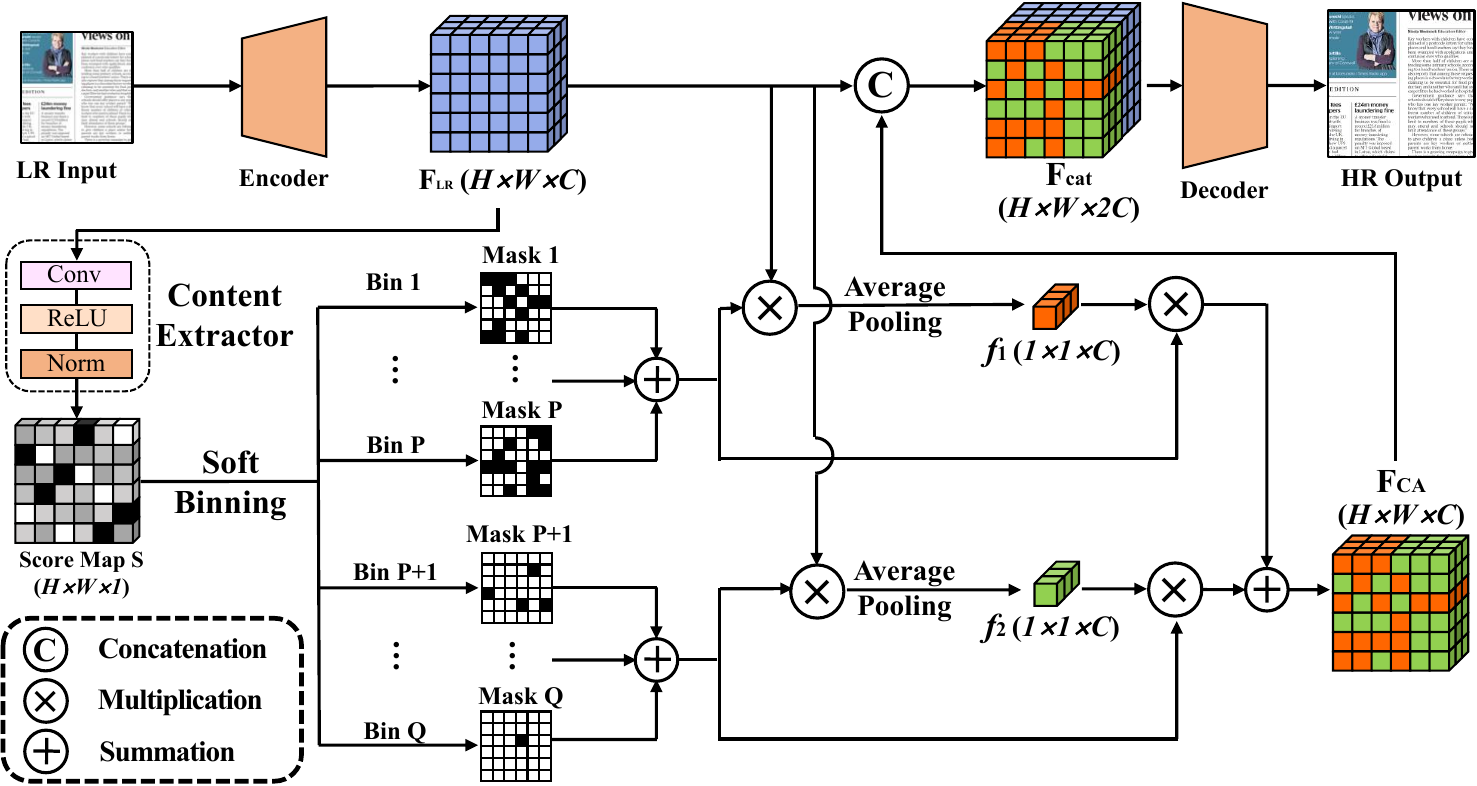}
  \caption{\label{fig3}The training pipeline of PixelSR. PixelSR uses a content extractor to generate a score map $S$, which classifies pixels into Q bins via soft pixel binning. Then, the first P bins and the rest Q-P bins are separately merged into two content-dependent groups. By aggregating pixel features within each group, we generate the content attention $F_{CA}$ and assign it to every pixel. In this way, PixelSR adopts a content-dependent and unlimited receptive field, improving SR performance efficiently. }
\end{figure*}

\section{Method}

In this section, we discuss how we achieve pixel classification in PixelSR to improve model performance and reduce inference time. We first present the pixel binning technique used during training. Then, we discuss how the pixel division in inference can reduce inference time without model performance loss. Finally, we introduce the network architecture. 

\subsection{Pixel Binning in Training}
It is acknowledged that the pixel distribution of screen content images is distinct from that of natural images. Previous works \cite{btc,yang2015perceptual} validate this statement in terms of the naturalness value. However, previous screen content SR methods \cite{itsrn,shen2022itsrn++,btc} do not treat these two types of content discriminatively, leaving a potential to boost the model performance of screen content SR. In addition, the receptive field is one of the most important factors in SR. However, in screen content SR, previous methods such as LIIF \cite{liif}, LTE \cite{lte} and BTC \cite{btc} are based on local implicit image function for continuous representation of images. With the use of feature unfolding, the receptive fields of these methods are limited to a small local region, which is the nearest $3\times3$ grid. To expand the receptive field, some methods \cite{ciaosr,crossscalenonlocal,gnn} compute the correlation between pixel features to involve more related pixels and inject non-local information. However, calculating correlation between dense pixel features is computationally heavy. Window partitioning could be required to conduct such heavy operation. However, window partitioning can lead to slow inference and window-limited \cite{gnn}. To achieve this goal efficiently, we introduce pixel binning to classify pixels into two types and construct their content attention, realizing a content-dependent and unlimited receptive field over the whole image, as demonstrated in Fig. \ref{fig3}. 

Specifically, we design a lightweight content extractor to output a two-dimensional score map given the input feature. The process of the content extractor can be expressed as: 
\begin{equation}
  S = Norm(ReLU(Conv(F_{LR}))),
  \label{eq1}
\end{equation}
\noindent where $F_{LR}$ is the feature from the encoder with the dimension of $H\times W \times C$ and $S$ is the normalized content score map with the dimension of $H\times W \times 1$. In other words, one convolutional layer with a kernel of 3 and ReLU activation generates a two-dimensional content score map. Then, we normalize the content score map along spatial dimensions $H$ and $W$ into the range of 0 to 1. 

We tend to partition pixels into content groups based on the normalized score map. However, hard binning breaks backpropagation and does not update the content extractor. Unlike hard binning which assigns each pixel to a single discrete bin with information loss, we design a soft binning technique to enable a differentiable and probabilistic assignment for each score to multiple bins. This allows the model to learn robust and content-aware representations while preserving gradient flow across bin boundaries. We define a set of bin centers $c_i$ uniformly spaced over the value range, written as:

\begin{equation}
  c_i = v_{min} + (i-1/2)\cdot \Delta,
  \label{eq2}
\end{equation}
where $i=1,\dots,N$ and $\Delta=\frac{v_{max}-v{min}}{N}$. Here, $v_{min}$ and $v_{max}$ are 0 and 1 respectively. For each score in the score map, we compute the soft assignment weight of each bin as: 

\begin{equation}
  w_i(x,y) = \frac{exp(-|S'(x,y)-c_i|/\tau)}{\sum_{j=1}^{N}exp(-|S'(x,y)-c_j|/\tau)},
  \label{eq3}
\end{equation}

\noindent where $\tau$ is a temperature hyperparameter to control the hardness of the assignment. A smaller temperature means a harder assignment. Thus, we set a low temperature value to make it closer to hard binning. After classifying pixels by the content extractor, another question is how to compute content attention from these classified pixels. It is not trivial because the pixels of each group are scattered over the whole image with a random quantity and random locations. Conventional CNN or MLP are not capable of achieving this goal because they are spatially sensitive. Here, we adopt a spatial averaging pooling operation, since the pooling operation is workable for random locations or any number of pixels in each content group. 

\begin{figure*}[t]
  \centering
  \label{fig2}
  \includegraphics[width=0.95\textwidth]{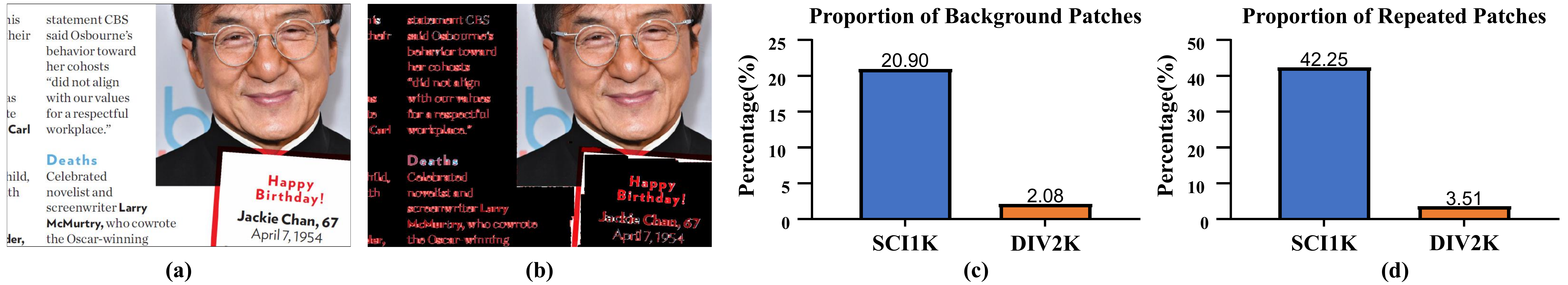}
  \caption{\label{fig2}(a) The example of screen content image; (b) the corresponding image where the background patches are masked out (in black) and the repeated patches are red highlighted. The rest patches are unique patches; (c) the proportions of the background patches in SCI1K (screen content image) and DIV2K (natural image) datasets respectively; (d) the proportions of the repeated patches (excluding background patches) in SCI1K (screen content image) and DIV2K (natural image) datasets respectively. }
\end{figure*}

One important factor affecting the performance of the soft binning and content attention is the number of bins. Since we tend to classify natural content and screen content only, one intuitive way is to set the number of bins to two. However, a small number of bins leads to too much discretization. It would prevent pixels from a smooth transition between bins and lead to overfitting. Therefore, we set more bins during training to prevent overfitting. Before conducting the spatial pooling operation, we merge multiple bins into two groups to represent the natural content and the screen content only. The group with the higher group centre is full of screen content pixels and the group with the lower group centre is filled with the rest of the pixels. One way to understand this behaviour is that the content extractor can be regarded as a learnable `edge' filter. Since the edges of screen content such as texts, charts, and tables are unnaturally sharp, the output (or the aforementioned `score') of these pixels tends to approach the higher group centre. Different from the conventional edge filter to extract continuous edge strength from the RGB input, the proposed content extractor is learnable in the feature space to discretize pixels into two content groups. As shown in Fig. \ref{fig3}, we generate the mask of two content groups which are merged from multiple bins. Then, we multiply the feature by the masks to get the features of each content group, and conduct a spatial average pooling operation to generate the mean feature of each group. Therefore, the mean feature of the first content group can be termed as:
\begin{equation}
  f_{1} = pooling(F_{LR}*\sum_{i=1}^PM_{i}),
  \label{eq2}
\end{equation}
\noindent where $M_i$ is the mask of one bin, and $f_1$ is the mean feature of the first content group with the dimension of $1 \times 1 \times C$. Similarly, the mean feature of the second content group is merged from the rest of bins, which is given as:
\begin{equation}
  f_{2} = pooling(F_{LR}*\sum_{i=P+1}^QM_{i}),
  \label{eq2}
\end{equation}
where $f_2$ is the mean feature of the second content group. To assign the final content attention to every pixel, we multiply each mean feature by the corresponding mask of its content group and merge them, which can defined as: 
\begin{equation}
  F_{CA} = f_{1}*\sum_{i=1}^PM_{i} + f_{2}*\sum_{i=P+1}^QM_{i},
  \label{eq3}
\end{equation}
\noindent where $F_{CA}$ is the content attention map with the dimension of $H \times W \times C$. The content attention map $F_{CA}$ then concatenates with the original feature $F_{LR}$ to get $F_{cat}$, which is further fed into the decoder to predict HR RGB values. 

\begin{figure*}[t]
  \centering
  \label{fig4}
  \includegraphics[width=\textwidth]{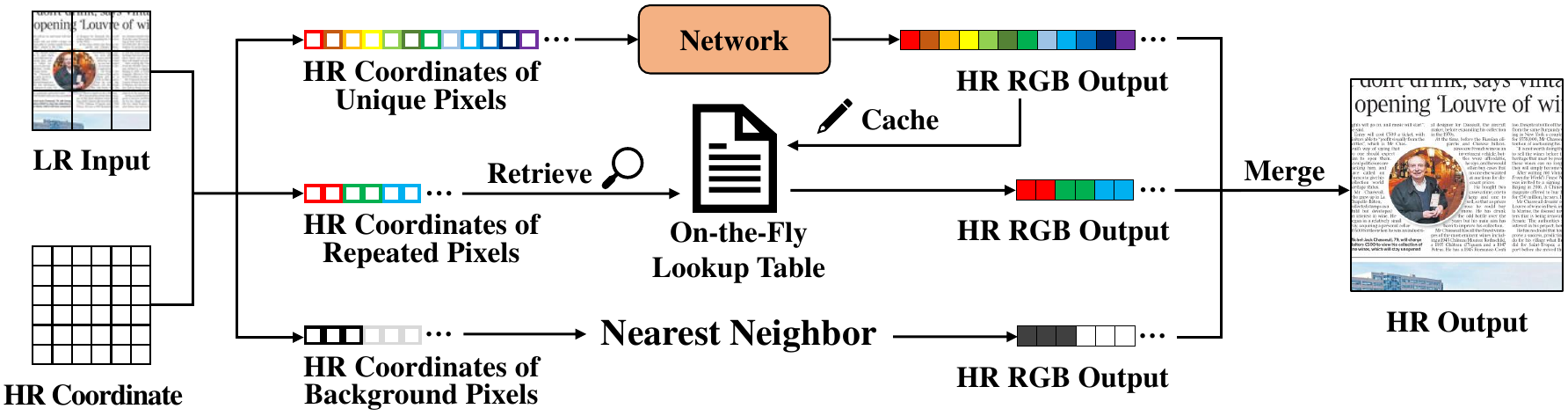}
  \caption{\label{fig4}The inference pipeline of PixelSR. For each test image, PixelSR divides pixels into unique pixels, repeated pixels, and background pixels. PixelSR first predicts the RGB outputs of unique pixels by network. The prediction results of unique pixels are cached in the on-the-fly lookup table. The RGB outputs of repeated pixels are directly retrieved from the cached on-the-fly lookup table, based on the inputs of RGB values of the nearest queried $3\times3$ LR patch and relative distance between the targeted HR pixel and the center pixel of the queried LR patch. Background pixels are predicted by using the nearest neighbor algorithm. After finishing the inference of one test image, we clean cached data in the on-the-fly lookup table, and repeat the procedure above for the next test image. }
\end{figure*}

\subsection{Pixel Division in Inference}
In addition to the unnatural sharpness of screen content, we suggest there are two more discrepancies between screen content images and natural images, which are high redundancy and high self-repetitiveness. To verify their differences, we conduct statistical comparisons between the screen content dataset SCI1K \cite{itsrn} and the natural image dataset DIV2K\cite{agustsson2017ntire}. 

\noindent \textbf{High Redundancy. }Although screen content images are informative, they are highly redundant because the background regions of screen content images are generally man-made. The pixel values of these regions are generally equal without any fluctuation. This is distinct from natural images. In natural images, backgrounds such as skies, out-of-focused regions, etc. are composed of low-frequency components but still fluctuate in pixel values. The possibility of the same RGB values is close to zero.  To verify this, we compare the proportions of background patches between SCI1K \cite{itsrn} and DIV2K \cite{agustsson2017ntire} datasets. Here, we densely divide images into $3\times3$ patches. The background patches are defined as the $3\times3$ patches where all pixel values are exactly equal. We take one SCI1K image as an example and mask out the background patches, which are demonstrated in Fig. \ref{fig2}(a) and (b) respectively. For statistical comparison, the proportion of background patches in the screen content dataset is
20.90\%, which is significantly higher than the one of the natural image DIV2K dataset (2.08\%), as demonstrated in Fig. \ref{fig2}(c). 

\noindent \textbf{High Self-Repetitiveness. }Screen content images contain a large quantity of vector graphics, which can be only made of simple shapes. Although screen content images can be largely different from each other, like using texts with different font sizes or font styles, the screen content image is highly likely to repeat itself for easy viewing. For example, in each screen content image, paragraphs are made of texts in the same style, tables or charts repeat in the same pattern, etc. To illustrate the high self-repetitiveness of the screen content image, we present repeated patches in the screen content image example by red highlighting in Fig. \ref{fig2}(b). Here, the repeated patch is defined as the $3\times3$ patch where it occurs more than once in this screen content image. From Fig. \ref{fig2}(b), we can see repeated patches frequently appear in the text region, and are unlikely to happen in the region of the natural image. We also conduct statistical comparisons of repeated patches between the SCI1K dataset and the DIV2K dataset. As shown in Fig. \ref{fig2}(d), the proportion of repeated patches in the screen content image dataset (42.25\% in the SCI1K dataset) is much higher than the proportion in the natural image dataset (3.51\% in the DIV2K dataset), after excluding background patches. 

From the analysis above, there exists high-redundancy and high self-repetitiveness in screen content images. These properties leave a potential to speed up inference in screen content SR. Specifically, we conduct pixel division to classify pixels into three types, which are unique pixels, repeated pixels, and background pixels. To classify them, we divide the LR input (in RGB values) into $3\times3$ patches with a stride of 1. Here, we define the HR pixel as the background pixel if the RGB values of the nine pixels in the nearest queried $3\times3$ LR patch are the same. After excluding background pixels. whether the pixel belongs to unique or repeated pixel depends on the RGB values of the nearest queried LR $3\times3$ patch and the relative distance between the targeted HR pixel and the center pixel of that LR patch. Unique pixels are the targeted HR pixels with the sets of the queried LR patches and the relative distances unique from each other. In contrast, repeated pixels have the sets of the $3\times3$ LR patches and the relative distances that have appeared in the unique pixels. In simple words, repeated pixels repeat the pixels in the category of unique pixels. We design the on-the-fly lookup table and utilize the nearest-neighbor algorithm to speed up the reconstruction of repeated pixels and background pixels, respectively. Therefore, the inference process is that we first use the nearest-neighbor algorithm to upscale the background pixels. Next, the network is used to process the unique pixels. After caching the unique pixels into the on-the-fly lookup table, the on-the-fly lookup table predicts the rest targeted pixels, which are repeated pixels. 

Different from conventional lookup tables in natural images, we design an on-the-fly lookup table where the cached data changes image by image. After network processing unique pixels, we cache RGB predictions with the inputs (including the RGB values of the nearest $3\times3$ LR patch and the relative distance) into the on-the-fly lookup table. We directly retrieve the RGB outputs from the on-the-fly lookup table for repeated pixels. After finishing the inference of one test image, we cleaned the on-the-fly lookup table. The on-the-fly lookup table is cached and retrieved again for the next test image. Thanks to the self-repetitiveness of screen content images, the proposed on-the-fly lookup table speeds up inference effectively. Compared to conventional lookup tables, the performance of the on-the-fly lookup table is not sensitive to the receptive field. We experimentally find $3\times3$ receptive field is sufficient for speeding up without PSNR loss. Also, the proposed on-the-fly lookup table is the first lookup table that can be used for scale arbitrary SR. 

\subsection{Network Details}
We present further details of the network design. For the content extractor, the input dimension, the output dimension, and the kernel size of the convolutional layer are 64, 1, and 3 respectively. The total number of bins for soft binning is 10. Then, the first bin with the lowest bin centre remains as the first content group, and the rest 9 bins with relatively high values of bin centres are merged as the second content group. The decoder in Fig. \ref{fig3} uses the BTC decoder \cite{btc}. 

\section{Experiment}

In this section, we demonstrate the effectiveness and efficiency of the proposed PixelSR by experiments. First, we introduce the experimental setup. Then, we present the quantitative and qualitative comparisons on SCI1K, SCID, and SIQAD datasets. Finally, detailed ablations with visualization are illustrated to investigate the PixelSR components. 

\begin{table*}[t!]
  \caption{Quantitative comparison with the EDSR backbone on the test set of the SCI1K (200 samples), SCID (40 samples), and SIQAD (22 samples) datasets in PSNR(dB). The best PSNR at each scale factor are bolded. }
  \label{table1}
  \centering
  \setlength{\tabcolsep}{3pt}
  \begin{tabular}{ ccc|ccc | c  ccccc}
    \toprule
    \multirow{2}{*}{Test set} & \multirow{2}{*}{Method} & \multirow{2}{*}{\#Params.} & \multicolumn{3}{c|}{In-training-scale} & \multicolumn{6}{c}{Out-of-training-scale} \\ 
     & & & $\times2$ & $\times3$ & $\times4$ & $\times5$ & $\times6$ & $\times7$ & $\times8$ & $\times9$ & $\times10$ \\
    \midrule
    
     \multirow{6}{*}{SCI1K} & Bicubic & - & 28.81 & 25.15 & 23.18 & 22.02 & 21.23 & 20.72 & 20.26 & 19.96 & 19.67 \\ 
     & LIIF\cite{liif} & 1.6M & 37.82 & 32.42 & 29.07 & 26.81 & 24.96 & 23.52 & 22.53 & 21.85 & 21.34 \\ 
     & ITSRN\cite{itsrn} & 1.9M & 37.42 & 32.31 & 28.98 & 26.72 & 24.88 & 23.47 & 22.48 & 21.81 & 21.32 \\
     & LTE\cite{lte} & 1.7M & 38.04 & 32.73 & 29.19 & 26.93 & 25.00 & 23.54 & 22.52 & 21.86 & 21.37 \\ 
     & BTC\cite{btc} & 1.7M & 38.02 & 32.80 & 29.19 & 26.89 & 25.05 & 23.60 & 22.58 & 21.90 & 21.39 \\ 
     & PixelSR (Ours) & 1.7M & \textbf{38.21} & \textbf{33.52} & \textbf{29.79} & \textbf{27.21} & \textbf{25.24} & \textbf{23.74} & \textbf{22.73} & \textbf{22.02} & \textbf{21.49} \\ 
     \midrule
     
     \multirow{6}{*}{SCID} & Bicubic & - &  25.22 & 22.78 & 21.60 & 20.90 & 20.42 & 20.04 & 19.77 & 19.51 & 19.29 \\ 
     & LIIF\cite{liif} & 1.6M & 33.13 & 27.86 & 24.85 & 23.09 & 22.01 & 21.36 & 20.92 & 20.56 & 20.29 \\ 
     & ITSRN\cite{itsrn} & 1.9M & 32.80 & 27.69 & 24.71 & 22.96 & 21.95 & 21.30 & 20.87 & 20.53 & 20.25 \\
     & LTE\cite{lte} & 1.7M & 33.38 & 28.08 & 25.02 & 23.18 & 22.09 & 21.40 & 20.95 & 20.59 & 20.30 \\ 
     & BTC\cite{btc} & 1.7M & 33.39 & 28.10 & 24.99 & 23.17 & 22.10 & 21.42 & 20.94 & 20.59 & 20.31 \\ 
     & PixelSR (Ours) & 1.7M & \textbf{33.58} & \textbf{28.34} & \textbf{25.18} & \textbf{23.29} & \textbf{22.15} & \textbf{21.48} & \textbf{21.00} &\textbf{20.67} & \textbf{20.35} \\ 
     \midrule
     
     \multirow{6}{*}{SIQAD} & Bicubic & - & 22.89 & 20.66 & 19.70 & 19.18 & 18.79 & 18.46 & 18.20 & 17.94 & 17.68 \\ 
     & LIIF\cite{liif} & 1.6M & 32.90 & 26.25 & 22.25 & 20.66 & 19.89 & 19.40 & 19.07 & 18.80 & 18.53 \\ 
     & ITSRN\cite{itsrn} & 1.9M & 32.51 & 26.05 & 22.18 & 20.59 & 19.83 & 19.34 & 19.00 & 18.75 & 18.47 \\
     & LTE\cite{lte} & 1.7M & 33.29 & 26.57 & 22.42 & 20.74 & 19.94 & 19.44 & 19.10 & 18.81 & 18.54 \\ 
     & BTC\cite{btc} & 1.7M & 33.22 & 26.65 & 22.42 & 20.77 & 19.96 & 19.44 & 19.10 & 18.83 & 18.55 \\ 
     & PixelSR (Ours) & 1.7M & \textbf{33.25} & \textbf{27.03} &\textbf{22.68} & \textbf{20.87} & \textbf{20.05} & \textbf{19.51} & \textbf{19.18} & \textbf{18.87} & \textbf{18.59} \\  
     \bottomrule
  \end{tabular}
\end{table*}
\begin{table*}[t!]
  \caption{Quantitative comparison with the SwinIRlight encoder in PSNR(dB). The best PSNR at each scale factor are bolded. }
  \label{table2}
  \centering
  \setlength{\tabcolsep}{3pt}
  \begin{tabular}{ ccc|ccc | c  ccccc}
    \toprule
    \multirow{2}{*}{Test set} & \multirow{2}{*}{Method} & \multirow{2}{*}{\#Params.} & \multicolumn{3}{c|}{In-training-scale} & \multicolumn{6}{c}{Out-of-training-scale} \\ 
     & & & $\times2$ & $\times3$ & $\times4$ & $\times5$ & $\times6$ & $\times7$ & $\times8$ & $\times9$ & $\times10$ \\
    \midrule
    
     \multirow{5}{*}{SCI1K} & LIIF\cite{liif} & 1.6M & 38.08 & 32.98 & 29.41 & 27.05 & 25.17 & 23.66 & 22.66 & 21.96 & 21.44 \\ 
     & ITSRN\cite{itsrn} & 1.9M & 37.82 & 32.64 & 29.13 & 26.94 & 25.07 & 23.63 & 22.65 & 21.95 & 21.44 \\
     & LTE\cite{lte} & 1.7M & 38.32 & 33.15 & 29.64 & 27.22 & 25.24 & 23.73 & 22.71 & 22.02 & 21.49 \\ 
     & BTC\cite{btc} & 1.7M & 38.36 & 33.35 & 29.70 & 27.28 & 25.30 & 23.78 & 22.75 & 22.03 & 21.49 \\ 
     & PixelSR (Ours) & 1.7M & \textbf{38.41} & \textbf{33.66} & \textbf{30.31} & \textbf{27.76} & \textbf{25.66} & \textbf{24.07} & \textbf{22.97} & \textbf{22.17} & \textbf{21.61} \\ 
     \midrule
     
     \multirow{5}{*}{SCID} & LIIF\cite{liif} & 1.6M & 33.37 & 27.97 & 24.93 & 23.18 & 22.14 & 21.41 & 20.94 & 20.62 & 20.30 \\
     & ITSRN\cite{itsrn} & 1.9M & 33.14 & 27.80 & 24.81 & 23.15 & 22.08 & 21.39 & 20.91 & 20.60 & 20.30 \\
     & LTE\cite{lte} & 1.7M & 33.38 & 28.08 & 25.02 & 23.18 & 22.09 & 21.40 & 20.95 & 20.59 & 20.30 \\ 
     & BTC\cite{btc} & 1.7M & 33.60 & 28.22 & 25.12 & 23.31 & 22.20 & 21.50 & 21.03 & 20.67 & 20.36 \\ 
     & PixelSR (Ours) & 1.7M & \textbf{33.81} & \textbf{28.59} & \textbf{25.36} & \textbf{23.48} & \textbf{22.32} & \textbf{21.58} & \textbf{21.08} &\textbf{20.73} & \textbf{20.42} \\ 
     \midrule
     
     \multirow{5}{*}{SIQAD} & LIIF\cite{liif} & 1.6M & 32.97 & 26.11 & 22.37 & 20.72 & 19.99 & 19.45 & 19.11 & 18.80 & 18.50 \\ 
     & ITSRN\cite{itsrn} & 1.9M & 32.56 & 25.89 & 22.29 & 20.77 & 19.99 & 19.44 & 19.07 & 18.74 & 18.45 \\
     & LTE\cite{lte} & 1.7M & 33.29 & 26.57 & 22.42 & 20.74 & 19.94 & 19.44 & 19.10 & 18.81 & 18.54 \\ 
     & BTC\cite{btc} & 1.7M & 33.20 & 26.40 & 22.50 & 20.83 & 20.09 & 19.53 & 19.16 & 18.84 & 18.54 \\ 
     & PixelSR (Ours) & 1.7M & \textbf{33.70} & \textbf{27.16} &\textbf{22.86} & \textbf{21.00} & \textbf{20.16} & \textbf{19.61} & \textbf{19.25} & \textbf{18.93} & \textbf{18.61} \\  
     \bottomrule
  \end{tabular}
\end{table*}

\subsection{Experimental Setup}

\textbf{Dataset. }We use SCI1K\cite{itsrn}, SCID\cite{ESIM}, and SIQAD\cite{yang2015perceptual} datasets for efficient screen content SR. We train one model on the train set of the SCI1K dataset, which has 800 data samples. The trained model is tested on the test set of the SCI1K, the SCID, and the SIQAD datasets, which have 200, 40, and 22 samples respectively. 

\noindent \textbf{Implementation Details.}
We use the SCI1K dataset \cite{itsrn} with the standard split for training and testing. We use the EDSR-baseline \cite{EDSR-baseline} and SwinIR-light \cite{swinir} as the backbone to extract features. In training, we downsample HR images by bicubic within a scale factor range $1\sim4$. We crop $48\times48$ LR patch as the input. The corresponding HR patch is converted to coordinate-RGB pairs \cite{liif}, and we randomly choose 2304 pairs for supervision. The LR and HR patches are augmented by flipping and rotating while training. L1 loss is used for training. We use an Adam \cite{Adam} optimizer to optimize the model. The initial learning rate is $0.0001$ and decays by half at every 200 epochs. The model is trained for 1000 epochs with a batch size of 16. The last epoch is used for testing. Experiments are conducted on one GeForce RTX 3090 GPU.

\begin{figure*}[t]
\newlength{\fsdurthree}
\setlength{\fsdurthree}{-3mm}
\centering
\begin{adjustbox}{valign=t}
\scriptsize
\begin{tabular}{ccccccc}
{\footnotesize Input} \hspace*{\fsdurthree}&
{\footnotesize LIIF~\cite{liif}} \hspace*{\fsdurthree}& 
{\footnotesize ITSRN~\cite{itsrn}} \hspace*{\fsdurthree}& 
{\footnotesize LTE~\cite{lte}}\hspace*{\fsdurthree}& 
{\footnotesize BTC~\cite{btc}}\hspace*{\fsdurthree}& 
{\footnotesize PixelSR}\hspace*{\fsdurthree}& 
{\footnotesize GT}\\

\includegraphics[height=0.136\textwidth]{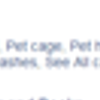}\hspace*{\fsdurthree} &
\includegraphics[height=0.136\textwidth]{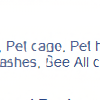}\hspace*{\fsdurthree} &
\includegraphics[height=0.136\textwidth]{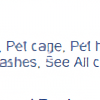}\hspace*{\fsdurthree} &
\includegraphics[height=0.136\textwidth]{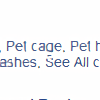}\hspace*{\fsdurthree} &
\includegraphics[height=0.136\textwidth]{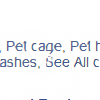}\hspace*{\fsdurthree} &
\includegraphics[height=0.136\textwidth]{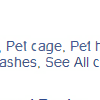}\hspace*{\fsdurthree} &
\includegraphics[height=0.136\textwidth]{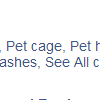}\hspace*{\fsdurthree}\\

\includegraphics[height=0.136\textwidth]{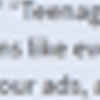}\hspace*{\fsdurthree} &
\includegraphics[height=0.136\textwidth]{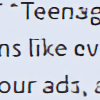}\hspace*{\fsdurthree} &
\includegraphics[height=0.136\textwidth]{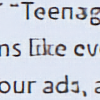}\hspace*{\fsdurthree} &
\includegraphics[height=0.136\textwidth]{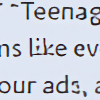}\hspace*{\fsdurthree} &
\includegraphics[height=0.136\textwidth]{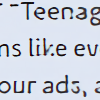}\hspace*{\fsdurthree} &
\includegraphics[height=0.136\textwidth]{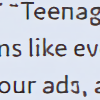}\hspace*{\fsdurthree} &
\includegraphics[height=0.136\textwidth]{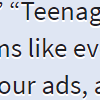}\hspace*{\fsdurthree}\\

\end{tabular}
\end{adjustbox}
\vskip 0.05cm
\caption{Qualitative comparison between different methods at the scale factors of 3 and 5.}
\label{fig:visual_all}
\vskip -0.2cm
\end{figure*}

\begin{figure*}[t]
\setlength{\fsdurthree}{-3mm}
\centering
\begin{adjustbox}{valign=t}
 \scriptsize
 \begin{tabular}{ccccccc}
 {\footnotesize Input} \hspace*{\fsdurthree}&
 {\footnotesize LIIF~\cite{liif}} \hspace*{\fsdurthree}& {\footnotesize ITSRN~\cite{itsrn}} \hspace*{\fsdurthree}& {\footnotesize LTE~\cite{lte}}\hspace*{\fsdurthree}& {\footnotesize BTC~\cite{btc}}\hspace*{\fsdurthree}& {\footnotesize PixelSR}\hspace*{\fsdurthree}& {\footnotesize GT}\\

\includegraphics[height=0.136\textwidth]{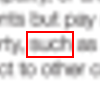}\hspace*{\fsdurthree} &
\includegraphics[height=0.136\textwidth]{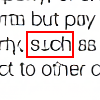}\hspace*{\fsdurthree} &
\includegraphics[height=0.136\textwidth]{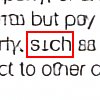}\hspace*{\fsdurthree} &
\includegraphics[height=0.136\textwidth]{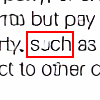}\hspace*{\fsdurthree} &
\includegraphics[height=0.136\textwidth]{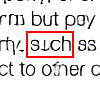}\hspace*{\fsdurthree} &
\includegraphics[height=0.136\textwidth]{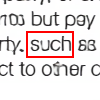}\hspace*{\fsdurthree} &
\includegraphics[height=0.136\textwidth]{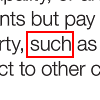}\hspace*{\fsdurthree}\\

{\footnotesize Pred.} \hspace*{\fsdurthree}&
{\footnotesize \textcolor{red}{cts}} \hspace*{\fsdurthree}& {\footnotesize s\textcolor{red}{i}ch} \hspace*{\fsdurthree}& {\footnotesize such}\hspace*{\fsdurthree}& {\footnotesize s\textcolor{red}{i}ch}\hspace*{\fsdurthree}& {\footnotesize such}\hspace*{\fsdurthree}& {\footnotesize such}\\
{\footnotesize {Conf.(\%)}} \hspace*{\fsdurthree}&
{\footnotesize 72.8} \hspace*{\fsdurthree}& {\footnotesize 99.7} \hspace*{\fsdurthree}& {\footnotesize 99.2}\hspace*{\fsdurthree}& {\footnotesize 96.2}\hspace*{\fsdurthree}& {\footnotesize \textcolor{red}{99.9}}\hspace*{\fsdurthree}& {\footnotesize }\\

\includegraphics[height=0.136\textwidth]{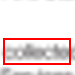}\hspace*{\fsdurthree} &
\includegraphics[height=0.136\textwidth]{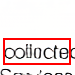}\hspace*{\fsdurthree} &
\includegraphics[height=0.136\textwidth]{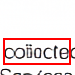}\hspace*{\fsdurthree} &
\includegraphics[height=0.136\textwidth]{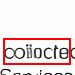}\hspace*{\fsdurthree} &
\includegraphics[height=0.136\textwidth]{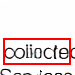}\hspace*{\fsdurthree} &
\includegraphics[height=0.136\textwidth]{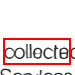}\hspace*{\fsdurthree} &
\includegraphics[height=0.136\textwidth]{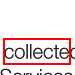}\hspace*{\fsdurthree}\\


{\footnotesize Pred.} \hspace*{\fsdurthree}&
{\footnotesize coll\textcolor{red}{o}cte} \hspace*{\fsdurthree}& 
{\footnotesize co\textcolor{red}{ijo}cte} \hspace*{\fsdurthree}& 
{\footnotesize co\textcolor{red}{i}l\textcolor{red}{o}cte} \hspace*{\fsdurthree}& 
{\footnotesize col\textcolor{red}{i}cte} \hspace*{\fsdurthree}& 
{\footnotesize collecte} \hspace*{\fsdurthree}& 
{\footnotesize collecte} \\
{\footnotesize {Conf.(\%)}} \hspace*{\fsdurthree}&
{\footnotesize 94.4} \hspace*{\fsdurthree}& 
{\footnotesize 92.4} \hspace*{\fsdurthree}& 
{\footnotesize 92.0} \hspace*{\fsdurthree}& 
{\footnotesize 98.0} \hspace*{\fsdurthree}& 
{\footnotesize \textcolor{red}{99.8}} \hspace*{\fsdurthree}& 
{\footnotesize }
 
 \end{tabular}
\end{adjustbox}
\vskip 0.05cm
\caption{Further qualitative comparison between methods for the scale factor of 4 with the scene text recognition (STR) results by a pretrained STR network \cite{bautista2022scene}. The wrong prediction of letters is coloured in red. The best confidence is coloured in red. }
\label{fig:visual_acc}
\vskip -0.2cm
\end{figure*}

\begin{figure*}[t!]
  \centering
  \label{figmask}
  \includegraphics[width=\textwidth]{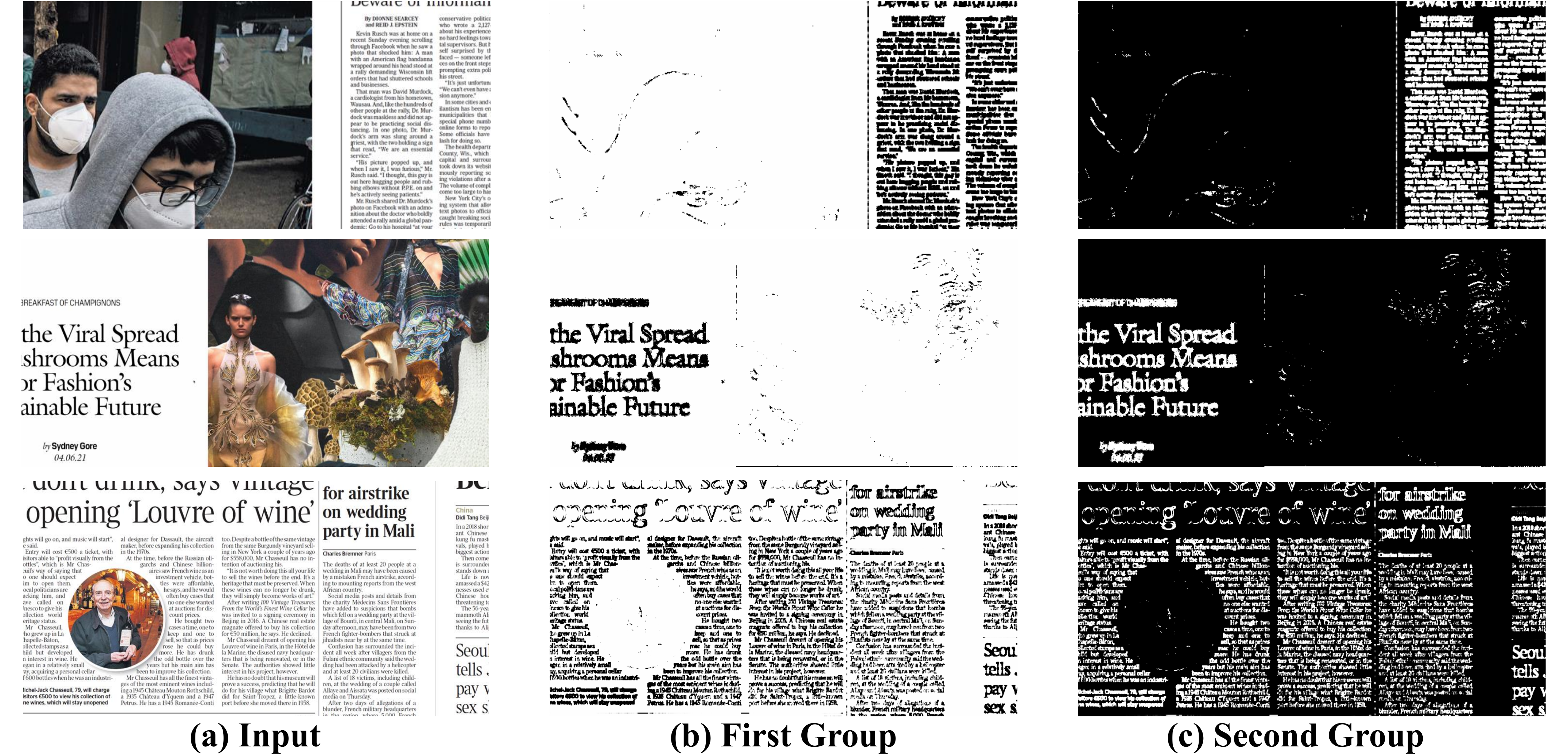}
  \caption{\label{figmask} Visualization of the pixel binning result for each input. The edges of screen content such as texts and charts are classified into the second group which has the higher value of bin centre. The rest of pixels are classified into the first group which has the lower value of bin centre. }
\end{figure*}

\begin{table*}[t]
  \caption{Analysis with the EDSR-backbone on the on-the-fly lookup table about the total inference time and PSNR on SCI1K test set (200 samples) in seconds and dB. The shorter inference time of each method at each scale factor is bolded. Inference time fluctuates within 1 second. }
  \label{table3}
  \setlength{\tabcolsep}{4pt}
  \centering
  \begin{tabular}{c | c | c| ccc|c  ccccc}
    \toprule
    \multirow{2}{*}{Metric}  & \multirow{2}{*}{Method}  & On-the-fly & \multicolumn{3}{c|}{In-training-scale} & \multicolumn{6}{c}{Out-of-training-scale} \\ 
     & &  lookup table & $\times2$ & $\times3$ & $\times4$ & $\times5$ & $\times6$ & $\times7$ & $\times8$ & $\times9$ & $\times10$ \\    
     
    \midrule
    \multirow{10}{*}{Inference time} & \multirow{2}{*}{LIIF\cite{liif}} & \xmark & 205 & 178 & 167 & 163 & 161 & 159 & 183 & 158 & 182 \\
    && \cmark & \textbf{142} & \textbf{128} & \textbf{123} & \textbf{124} & \textbf{149} & \textbf{135} & \textbf{153} & \textbf{130} & \textbf{169} \\
    \cmidrule{2-12}
    &\multirow{2}{*}{ITSRN\cite{itsrn}} & \xmark & 279 & 262 & 252 & 252 & 251 & 247 & 243 & 247 & 249 \\
    && \cmark & \textbf{193} & \textbf{187} & \textbf{188} & \textbf{191} & \textbf{200} & \textbf{197} & \textbf{207} & \textbf{205} & \textbf{213} \\
    \cmidrule{2-12}
    &\multirow{2}{*}{LTE\cite{lte}} & \xmark & 199 & 188 & 184 & 182 & 215 & 180 & 214 & 179 & 214 \\
    && \cmark & \textbf{142} & \textbf{136} & \textbf{137} & \textbf{140} & \textbf{172} & \textbf{145} & \textbf{178} & \textbf{151} & \textbf{183} \\
    \cmidrule{2-12}
    &\multirow{2}{*}{BTC\cite{btc}} & \xmark & 176 & 165 & 161 & 160 & 186 & 158 & 185 & 157 & 185 \\
    && \cmark & \textbf{122} & \textbf{119} & \textbf{130} & \textbf{132} & \textbf{148} & \textbf{127} & \textbf{153} & \textbf{132} & \textbf{159} \\
    \cmidrule{2-12}
    &\multirow{2}{*}{PixelSR (Ours)} & \xmark & 227 & 183 & 170 & 163 & 179 & 159 & 177 & 158 & 178 \\
    & & \cmark & \textbf{149} & \textbf{130} & \textbf{125} & \textbf{124} & \textbf{142} & \textbf{127} & \textbf{147} & \textbf{131} & \textbf{152} \\
    
    \midrule
    \multirow{10}{*}{PSNR} & \multirow{2}{*}{LIIF\cite{liif}} & \xmark & 37.82 & 32.42 & 29.07 & 26.81 & 24.96 & 23.52 & 22.53 & 21.85 & 21.34 \\
    && \cmark & 37.83 & 32.43 & 29.07 & 26.81 & 24.96 & 23.52 & 22.53 & 21.85 & 21.34 \\
    \cmidrule{2-12}
    &\multirow{2}{*}{ITSRN\cite{itsrn}} & \xmark & 37.42 & 32.31 & 28.98 & 26.52 & 24.63 & 23.25 & 22.30 & 21.68 & 21.20 \\
    && \cmark & 37.42 & 32.32 & 28.98 & 26.52 & 24.63 & 23.25 & 22.30 & 21.68 & 21.20 \\
    \cmidrule{2-12}
    &\multirow{2}{*}{LTE\cite{lte}} & \xmark & 38.04 & 32.73 & 29.19 & 26.93 & 25.00 & 23.54 & 22.52 & 21.86 & 21.37  \\
    && \cmark & 38.07 & 32.74 & 29.20 & 26.93 & 25.00 & 23.54 & 22.53 & 21.86 & 21.37 \\
    \cmidrule{2-12}
    &\multirow{2}{*}{BTC\cite{btc}} & \xmark & 38.02 & 32.80 & 29.19 & 26.89 & 25.05 & 23.60 & 22.58 & 21.90 & 21.39 \\
    && \cmark & 38.02 & 32.80 & 29.19 & 26.90 & 25.05 & 23.61 & 22.58 & 21.90 & 21.39 \\
    \cmidrule{2-12}
    & \multirow{2}{*}{PixelSR (Ours)} & \xmark & 38.21 & 33.50 & 29.78 & 27.20 & 25.24 & 23.74 & 22.73 & 22.03 & 21.49 \\
    & & \cmark & 38.21 & 33.50 & 29.79 & 27.21 & 25.24 & 23.74 & 22.73 & 22.02 & 21.49 \\
    
    \bottomrule
  \end{tabular}
\end{table*}

\begin{figure}[t]
\setlength{\fsdurthree}{-3.0mm}
\centering
\begin{adjustbox}{valign=t}
 \scriptsize
 \begin{tabular}{ccccccc}

 \includegraphics[height=0.13\textwidth]{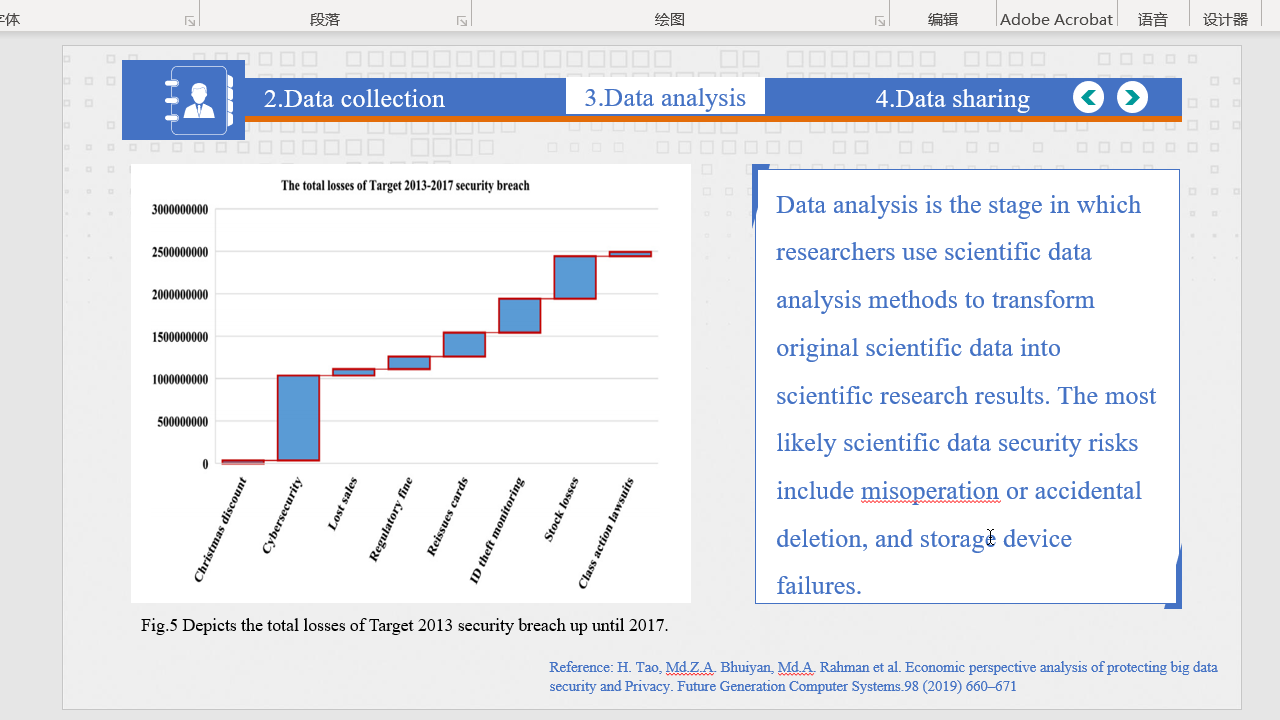}\hspace*{\fsdurthree} &
\includegraphics[height=0.13\textwidth]{00002.png}\hspace*{\fsdurthree} \\

{\footnotesize SCI1K 002 ($\times2$)} \hspace*{\fsdurthree}&
 {\footnotesize SCI1K 002 ($\times4$)} \hspace*{\fsdurthree} \\

{\footnotesize Time(wo/w): 0.556s / 0.245s} \hspace*{\fsdurthree}&
 {\footnotesize Time(wo/w): 0.422s / 0.257s} \hspace*{\fsdurthree} \\ 

{\footnotesize Mem.(wo/w): 1306M/1380M} \hspace*{\fsdurthree}&
 {\footnotesize Mem.(wo/w): 680M/770M} \hspace*{\fsdurthree} \\

\includegraphics[height=0.13\textwidth]{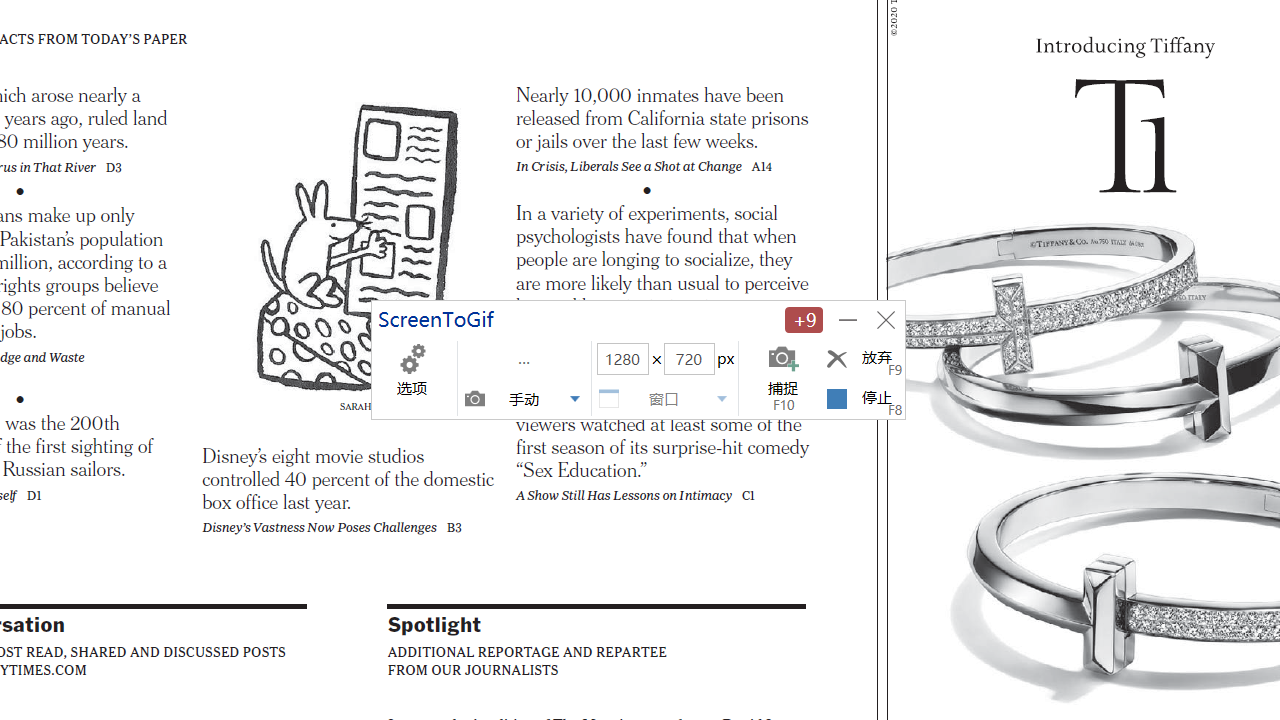}\hspace*{\fsdurthree}&
\includegraphics[height=0.13\textwidth]{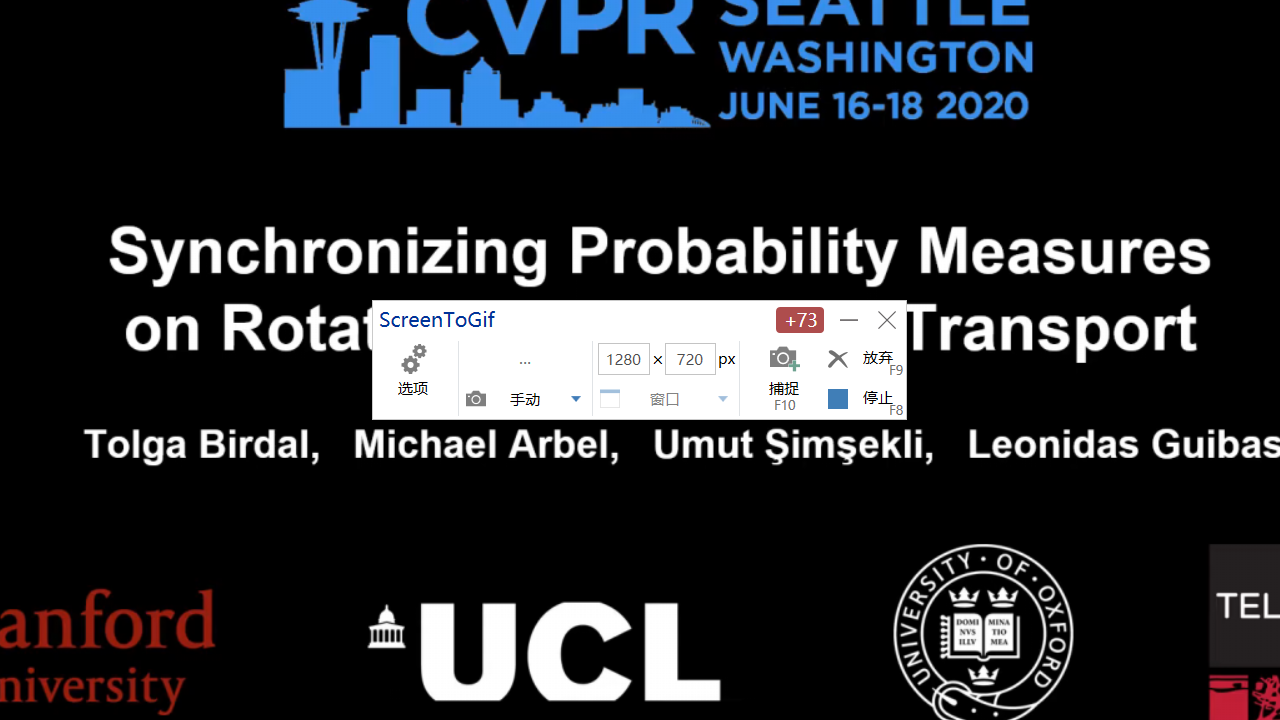}\hspace*{\fsdurthree} 
\\

 {\footnotesize SCI1K 009 ($\times2$)} \hspace*{\fsdurthree}& {\footnotesize SCI1K 072 ($\times2$)}\hspace*{\fsdurthree}\\

 {\footnotesize Time(wo/w): 0.554s / 0.289s} \hspace*{\fsdurthree}& {\footnotesize Time(wo/w): 0.551s / 0.338s}\hspace*{\fsdurthree}\\

  {\footnotesize Mem.(wo/w): 1304M/1356M} \hspace*{\fsdurthree}& {\footnotesize Mem.(wo/w): 1304M / 1356M}\hspace*{\fsdurthree}\\

 \end{tabular}
\end{adjustbox}
\vskip 0.05cm
\caption{Comparison of inference time (excluding the running time of the backbone encoder) and memory cost for PixelSR without and with the on-the-fly lookup table on image samples, which are mainly composed of screen content such as tables, charts, texts, etc. It demonstrates that the proposed on-the-fly lookup table is particularly effective on the SR of screen content. }
\label{fig:comp}
\vskip -0.2cm
\end{figure}

\subsection{Evaluation}

\noindent \textbf{Quantitative Results.} To evaluate the effectiveness and efficiency of the proposed PixelSR, we conduct extensive quantitative comparisons against the state-of-the-art screen content super-resolution methods, including LIIF \cite{liif}, ITSRN \cite{itsrn}, LTE \cite{lte}, and BTC \cite{btc}, with two different backbones, including EDSR \cite{EDSR-baseline} and SwinIR-light \cite{swinir}, on three benchmark datasets (SCI1K, SCID, and SIQAD). All methods are evaluated at the in-training scales ($\times2$, $\times3$, $\times4$) and the out-of-training scales (ranging from $\times5$ to $\times10$). PSNR in dB is taken as the evaluation metric. The results of the EDSR-baseline backbone and the SwinIR-light backbone are reported in TABLE \ref{table1} and TABLE \ref{table2}, respectively. We can observe that PixelSR consistently achieves the highest PSNR across all test sets at all scale factors, validating a clear and robust performance advantage. For the in-training scales under the EDSR backbone in TABLE \ref{table1}, PixelSR on the SCI1K dataset outperforms the state-of-the-art performance by a noticeable margin. For example, the metric gain reaches 0.72 dB at the scale factor of 3. For the out-of-training scales, PixelSR still maintains a leading PSNR, which is higher than all competing methods at all testing scales. On the SCID and SIQAD datasets, PixelSR also ranks first across all scales. For instance, on SIQAD at ×3 scale, PixelSR reaches 27.03 dB, surpassing the second-best BTC (which is 26.65 dB) by 0.38 dB. In addition, on the SCID benchmark at the scale factor of 3, PixelSR reaches 28.34dB, surpassing the state-of-the-art model by 0.24dB. When equipped with the SwinIR-light backbone, PixelSR still consistently achieves the best results, as illustrated in TABLE \ref{table2}. On the SCI1K dataset at a scale factor of 4, PixelSR reaches 30.31 dB, outperforming the second-best model BTC (29.70 dB) by a substantial margin of 0.61 dB. At the scale factor of 5, which is out-of-training, PixelSR again achieves the highest PSNR of 27.76 dB with a metric gain of 0.48dB compared to the second-best BTC model. On the SCID and SIQAD benchmark datasets, PixelSR still ranks first at all the scale factors. For example, PixelSR achieves 28.54 dB on SCID at ×3 scale. Compared to 28.22 dB of BTC, the metric gain reaches 0.32dB. It is worth noting that all compared methods, including PixelSR, have a comparable number of parameters (approximately 1.6M–1.9M), so the performance gains are not due to increased model capacity but due to the model design of PixelSR. In summary, regardless of the backbone architecture, PixelSR presents superior and efficient reconstruction accuracy at both the in-training and out-of-training scales. Also, it generalizes well to other screen content benchmarks. 

\begin{figure*}[t!]
  \centering
  
  \includegraphics[width=\textwidth]{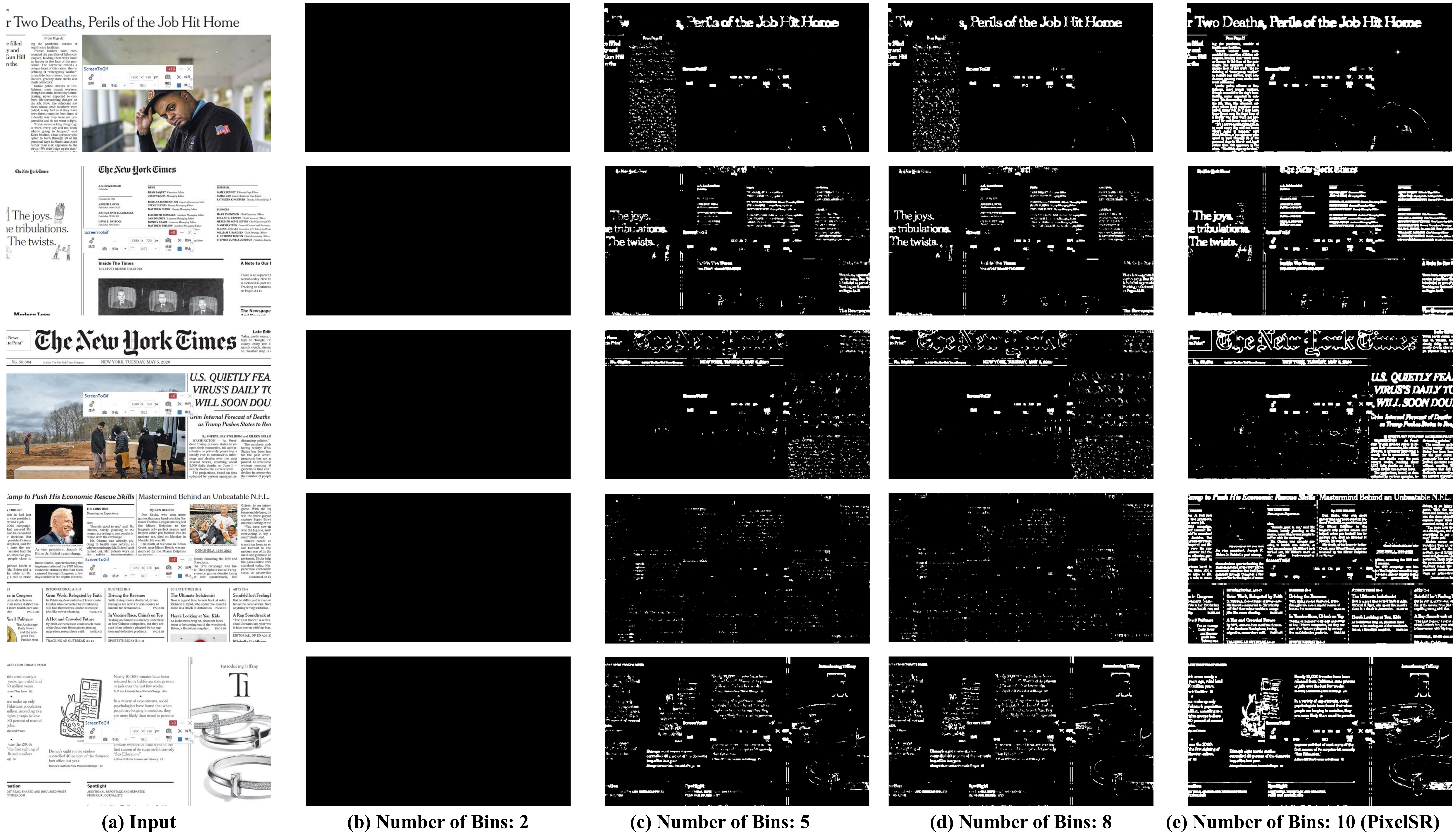}
  \caption{\label{figmaskbin}Visual Comparison of the pixels classified into the second group for the number of bins of (b) 2, (c) 5, (d) 8, and (e) 10 (PixelSR) during training. It shows that an insufficient number of bins, such as two, leads to overfitting, and a large number of bins, such as 10 in PixelSR, tend to provide satisfactory binning results. }
\end{figure*}

\begin{table*}[t!]
  \caption{PSNR comparison on the number of bins in SCI1K test set. The highest PSNR at each scale factor is bolded. }
  \label{table4}
  \centering
  \setlength{\tabcolsep}{7pt}
  \begin{tabular}{ c|ccc | c  ccccc}
    \toprule
    \multirow{2}{*}{Number of bins} & \multicolumn{3}{c|}{In-training-scale} & \multicolumn{6}{c}{Out-of-training-scale} \\ 
      & $\times2$ & $\times3$ & $\times4$ & $\times5$ & $\times6$ & $\times7$ & $\times8$ & $\times9$ & $\times10$ \\
    \midrule
     0 & 38.02 & 32.80 & 29.19 & 26.89 & 25.05 & 23.60 & 22.58 & 21.90 & 21.39 \\
     2 & 37.32 & 33.09 & 29.67 & 27.16 & 25.19 & 23.68 & 22.67 & 21.96 & 21.45 \\ 
     5 & 37.70 & 32.90 & 29.56 & 27.16 & 25.26 & 23.77 & 22.72 & 22.00 & 21.47 \\ 
     8 & 37.90 & 33.09 & 29.64 & 27.16 & 25.16 & 23.65 & 22.64 & 21.97 & 21.47 \\ 
     10 (PixelSR) & \textbf{38.21} & \textbf{33.52} & \textbf{29.79} & \textbf{27.21} & \textbf{25.24} & \textbf{23.74} & \textbf{22.73} & \textbf{22.02} & \textbf{21.49} \\ 
     \bottomrule
  \end{tabular}
\end{table*}

As for the inference time, we present a comparison of the inference time between different methods \cite{liif,itsrn,lte,btc} with the EDSR-baseline backbone in TABLE \ref{table3}. Since the effect of the on-the-fly lookup table is dependent on the image content, we compare the total inference time on the whole SCI1K test set (200 samples) in seconds. By comparing LIIF without the on-the-fly lookup table, ITSRN without the on-the-fly lookup table, LTE without the on-the-fly lookup table, BTC without the on-the-fly lookup table, and PixelSR with the on-the-fly lookup table, we can observe that PixelSR takes the shortest inference time on SCI1K test set among the competitors.

\noindent \textbf{Qualitative Results.} We present a qualitative comparison between different methods in Fig. \ref{fig:visual_all}. Overall, PixelSR consistently produces sharper edges and better structures in screen content SR compared to other approaches including LIIF \cite{liif}, ITSRN \cite{itsrn}, LTE \cite{lte},  and BTC \cite{btc}. In the first group of examples, PixelSR reconstructs the word "cage" and "See" with the finest level of detail, preserving character structures that are blurred or distorted in competing results. In the second group, when reconstructing the word "like", PixelSR stands out by reconstructing the letters with the most complete and coherent shape, while other methods show noticeable degradation or incomplete forms. Further qualitative comparison with the scene text recognition is illustrated in Fig. \ref{fig:visual_acc}, showing that the prediction results of the PixelSR model are best recognized. These visual comparisons proves PixelSR’s superior capability in preserving semantic details and structural integrity in screen content SR.

\begin{table*}[t!]
  \caption{PSNR comparison on the number of layers in the content extractor in SCI1K test set with the EDSR-baseline backbone. The highest PSNR at each scale factor is bolded. }
  \label{table5}
  \centering
  \setlength{\tabcolsep}{7pt}
  \begin{tabular}{ c|ccc | c  ccccc}
    \toprule
    \multirow{2}{*}{Number of layers} & \multicolumn{3}{c|}{In-training-scale} & \multicolumn{6}{c}{Out-of-training-scale} \\ 
      & $\times2$ & $\times3$ & $\times4$ & $\times5$ & $\times6$ & $\times7$ & $\times8$ & $\times9$ & $\times10$ \\
    \midrule
    1 (PixelSR) & 38.21 & \textbf{33.52} & \textbf{29.79} & 27.21 & \textbf{25.24} & \textbf{23.74} & \textbf{22.73} & \textbf{22.02} & \textbf{21.49} \\ 
     2 & 37.77 & 33.05 & 29.70 & \textbf{27.24} & 25.21 & 23.71 & 22.67 & 21.99 & 21.46 \\
     3 & \textbf{38.22} & 33.29 & 29.60 & 27.17 & 25.18 & 23.64 & 22.60 & 21.93 & 21.46 \\ 
     
     \bottomrule
  \end{tabular}
\end{table*}

\begin{figure*}[t!]
  \centering
  \includegraphics[width=\textwidth]{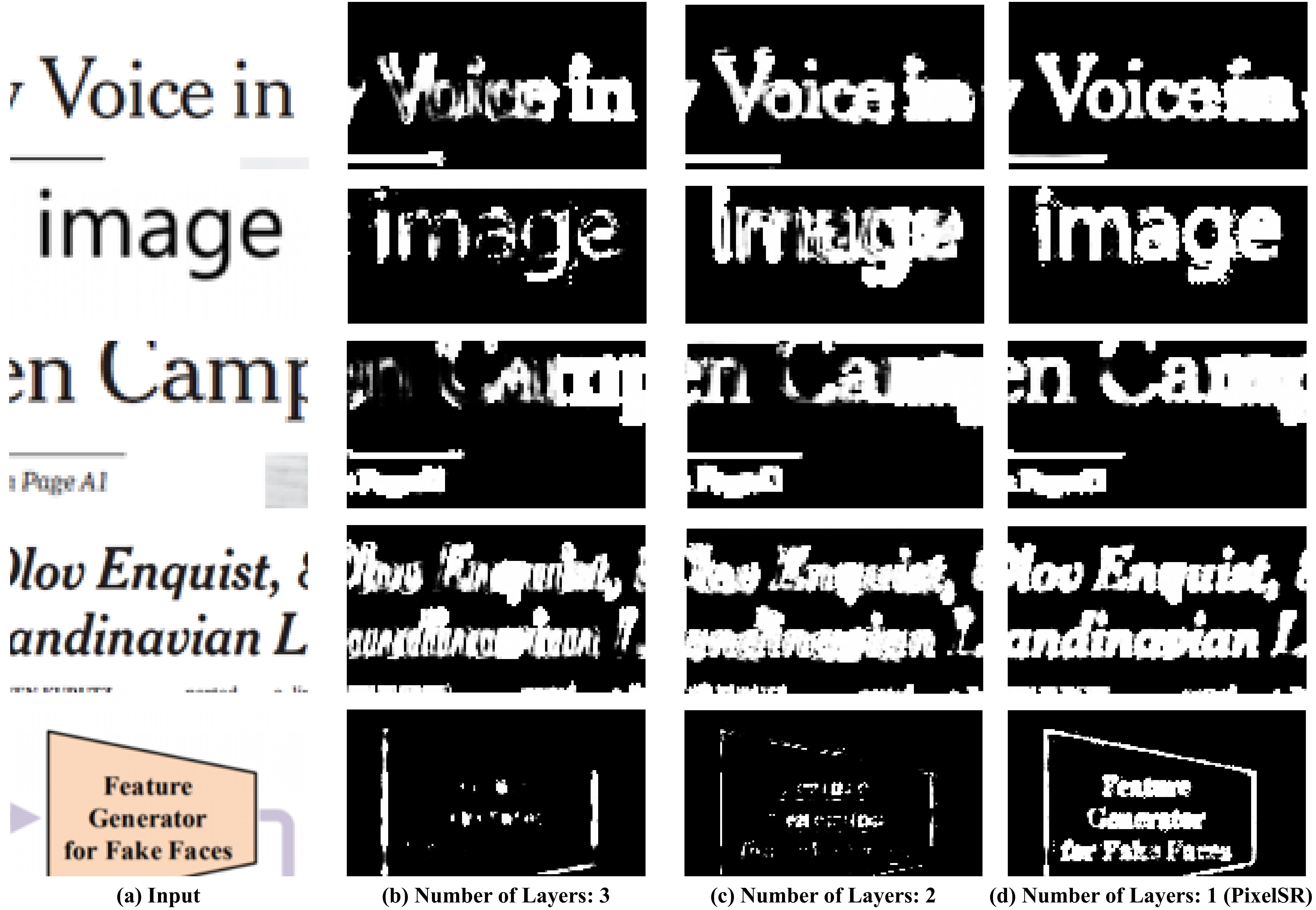}
  \caption{\label{figmasklayer}Visual Comparison of the pixels classified into the second group for the number of layers of (b) 3, (c) 2, and (d) 1 (PixelSR) during training. It shows that one layer of convolution provides more satisfactory binning results than more layers in PixelSR. }
\end{figure*}

\subsection{Visualization of Binning Result}

Here, we present the visualization of binning results in Fig. \ref{figmask}. As shown in Fig. \ref{figmask}, pixels in plain areas are generally classified into the first content group and pixels at the edges of screen content are classified into the second content group. The second group has a higher center value than the center value of the first group. In addition, pixels in natural images are rarely classified into the second group. It validates the fact that the content extractor can be regarded as a learnable `edge' filter. Since the edges of screen content are unnaturally sharp, these pixels are likely to be classified into the bins with the higher bin center. By taking the mean feature among these pixels in these two groups, we extract the mean representation of the natural content and screen content separately. After concatenating them back to the original feature, we assign a content-dependent attention which covers the whole image to every pixel. 

\subsection{Analysis of the On-the-fly Lookup Table}
In this section, we present a detailed analysis of the effect of the on-the-fly lookup table. Since the on-the-fly lookup table is specifically designed for screen content images but not natural images, we compare the inference time and memory cost of several screen content examples which are mainly composed of tables, charts, and texts, as illustrated in Fig. \ref{fig:comp}. Since the on-the-fly lookup table focuses on speeding up the decoder part, we exclude the running time of the backbone encoder and then compare the inference time for the PixelSR method without and with the on-the-fly lookup table. From Fig. \ref{fig:comp}, we can see that the on-the-fly lookup table can reduce the inference time significantly. For instance, the reduction of inference time reaches 55.9\% for the image sample of SCI1K 002. In addition, the success of the on-the-fly lookup table relies on the self-repetitiveness of screen content images. Therefore, it only needs to cache the RGB inputs, coordinates, and RGB outputs of the unique pixels in each test image. In this way, the memory cost of the on-the-fly lookup table is small, as demonstrated in Fig. \ref{fig:comp}. 

To demonstrate that the on-the-fly lookup can work on various screen content SR baselines, TABLE \ref{table3} compares the total inference time and PSNR without and with the on-the-fly lookup table for LIIF \cite{liif}, ITSRN \cite{itsrn}, LTE \cite{lte}, BTC \cite{btc} and PixelSR. Because the effect of the on-the-fly table varies with image content, we compare the total inference time for the whole SCI1K test set to show it works for general screen content images. As shown in TABLE \ref{table3}, the inference times of all screen content SR methods at all scale factors are reduced with the presence of the on-the-fly lookup table. Also, the PSNR of all methods at all scale factors just fluctuates with a small value of PSNR. This is because the functionality of the on-the-fly lookup table is rooted in the properties of screen content images. These experiments validate that the on-the-fly lookup table can speed up the model inference in screen content SR without performance loss. 

\subsection{Ablation}
In this section, we carry out an ablation study on the pixel binning during training in PixelSR. 

\noindent \textbf{The effect of the number of bins. }The number of bins during training is an important factor to determine the performance of the proposed pixel binning. To ensure a smooth transition between bins, we set a large number of bins during training. Specifically, we compare the experimental results of the number of bins between 2, 5, 8, and 10. In Fig. \ref{figmaskbin}, we present a qualitative comparison of the binning result of the second group. We can see that with a larger number of bins during training, the binning result becomes finer and finer. For only two bins during training, the binning result simply overfits and fails. TABLE \ref{table4} further compares the PSNR results between different number of bins, showing that a large number of bins (which is 10 in PixelSR) leads to the best metric results. 

\noindent \textbf{The effect of the number of layers. }We further investigate the effect of the number of convolutional layers in the content extractor on the binning results. Since the decoder of PixelSR relies on feature unfolding \cite{liif} to extract the nearest $3\times3$ patch feature, we suggest one convolutional layer with a kernel size of 3 is sufficient. TABLE \ref{table5} compares the PSNR results between different numbers of convolutional layers, including 1 (PixelSR), 2, and 3. According to the experimental results, we can see that more convolutional layers do not lead to an extra advantage on the model performance. Also, Fig. \ref{figmasklayer} illustrates the regions where the binning results of the numbers of layers of 2 and 3 are worse than just one convolutional layer. This is because more convolutional layer equivalently expands the receptive field to be larger than $3\times3$. In this case, the receptive field of the content extractor does not coincide with the major receptive field of the decoder (which is $3\times3$ by feature unfolding). Therefore, injecting the pixels which are not seen by the decoder into the content extractor may introduce noise and lead to worse binning results. 


\section{Conclusion}
In summary, we propose PixelSR, improving the training and inference by pixel classification for efficient screen content SR. In training, PixelSR classifies pixels into content-dependent groups via soft pixel binning. By computing content attention from pixel groups, PixelSR introduces content-dependent and non-local information over the whole image to the implicit neural representation. For the inference of each test image, PixelSR divides targeted HR pixels into unique pixels, repeated pixels, and background pixels. After network processing unique pixels, we cache their prediction results into the on-the-fly lookup table. The inferences of repeated pixels and background pixels are fastened by the on-the-fly lookup table and the nearest neighbor algorithm, without the cost of performance drop. Experiments show that our PixelSR model achieves state-of-the-art performance with the shortest inference time. As for the limitation, since the on-the-fly lookup table relies on the property of screen content images, it would not function well in natural images.

\bibliographystyle{IEEEtran}
\bibliography{main}

\end{document}